\documentclass[10pt,twocolumn,letterpaper]{article}

\usepackage{cvpr}              

\usepackage[sort, numbers]{natbib}

\usepackage[accsupp]{axessibility}  
\usepackage{graphicx}
\usepackage{amsmath}
\usepackage{amssymb}
\usepackage{booktabs}
\usepackage[ruled,noend]{algorithm2e}
\newcommand{\algcomment}[1]{\hfill$\triangleright$\textcolor{mydarkblue}{ \texttt{#1}}}
\usepackage{wrapfig}
\usepackage{multirow}
\usepackage{bbm}
\usepackage{graphics}
\usepackage{paralist}
\usepackage{color, colortbl}
\usepackage{xcolor}
\definecolor{mydarkblue}{rgb}{0,0.08,0.45}
\definecolor{magenta}{rgb}{1.0, 0.0, 1.0}

\usepackage[pagebackref,colorlinks,bookmarks=false,citecolor=blue,linkcolor=blue,urlcolor=blue]{hyperref}

\usepackage[capitalize]{cleveref}
\crefname{section}{Sec.}{Secs.}
\Crefname{section}{Section}{Sections}
\Crefname{table}{Table}{Tables}
\crefname{table}{Tab.}{Tabs.}

\usepackage{amsmath,amsfonts,bm}

\def\eqref#1{equation~\ref{#1}}

\def\1{\bm{1}}

\def\vx{{\bm{x}}}

\def\vz{{\bm{z}}}

\DeclareMathAlphabet{\mathsfit}{\encodingdefault}{\sfdefault}{m}{sl}
\SetMathAlphabet{\mathsfit}{bold}{\encodingdefault}{\sfdefault}{bx}{n}

\newcommand{\Var}{\mathrm{Var}}

\DeclareMathOperator*{\argmax}{arg\,max}

\def\cvprPaperID{8269} 
\def\confName{CVPR}
\def\confYear{2023}

\begin{document}

\title{Improving Zero-shot Generalization and Robustness of Multi-modal Models}
\author{Yunhao Ge$^{1,2*}$,  Jie Ren$^{1*}$,  Andrew Gallagher$^{1}$, Yuxiao Wang$^{1}$, Ming-Hsuan Yang$^{1}$, \\  Hartwig Adam$^{1}$, Laurent Itti$^{2}$, Balaji Lakshminarayanan$^{1\dagger}$, Jiaping Zhao$^{1\dagger}$ \\
$^{1}$Google Research \ \ \  $^{2}$University of Southern California \\  
{\tt \small
$^{*}$co-first author, \
$\dagger$correspondence to \{balajiln, jiapingz\}@google.com} 
}
\maketitle

\begin{abstract}
Multi-modal image-text models such as CLIP and LiT have demonstrated impressive performance on image classification benchmarks 
and their zero-shot generalization ability is particularly exciting. 
While the top-5 zero-shot accuracies of these models are very high, the top-1 accuracies are much lower (over 25\% gap in some cases). We investigate the reasons for this performance gap and 
find that many of the failure cases are caused by ambiguity in the text prompts. 
First, we develop a simple and efficient zero-shot post-hoc  method to identify images whose top-1 prediction is likely to be incorrect, by measuring consistency of the predictions w.r.t. multiple prompts and image transformations. 
We show that our procedure better predicts mistakes, outperforming the popular max logit baseline on selective prediction tasks. 
Next, we propose a simple and efficient way to improve accuracy on such uncertain images by making use of the WordNet hierarchy; specifically we augment the original class by incorporating its parent and children from the semantic label hierarchy, and plug the augmentation into text prompts.  
 We conduct experiments on both CLIP and LiT models with five different ImageNet-based datasets. For CLIP, our method \textbf{improves the top-1 accuracy by 17.13\% on the uncertain subset and  3.6\% on the entire ImageNet validation set.} We also show that our method improves across ImageNet shifted datasets, four other datasets, and other model architectures such as LiT.
\textbf{The proposed method\footnote{Work carried out mainly at Google} is hyperparameter-free, requires no additional model training and can be easily scaled to other large multi-modal architectures.} Code is available at \small{\textcolor{magenta}{\url{https://github.com/gyhandy/Hierarchy-CLIP}}}.

\end{abstract}
\vspace{-15pt}
\section{Introduction}
Vision-language multi-modal models trained on large-scale data have achieved significant success in numerous domains and have demonstrated excellent zero-shot generalization ability \citep{radford2021learning,zhai2022lit,pham2021combined,jia2021scaling,ramesh2021zero,ge2022dall}.
%
Given a test image and a set of candidate class labels, one can compute the similarity between the embedding of the image and the embedding of each candidate class labels, and predict the class as the one with the highest similarity. 
The zero-shot top-$1$ accuracy for ImageNet \cite{deng2009imagenet} using CLIP variants (CLIP ViT-L) matches the performance of the original ResNet model trained from scratch. 
Recently, CLIP has been found to be more robust to distribution shift than ResNet, achieving good performance on ImageNet-V2 \citep{recht2019imagenet}, ImageNet-R \citep{hendrycks2021many}, ImageNet-A \citep{hendrycks2021natural}, and ImageNet-Sketch \citep{wang2019learning}.

\begin{figure*}[ht]
\vspace{-15pt}
\centering
\includegraphics[width=0.95\textwidth]{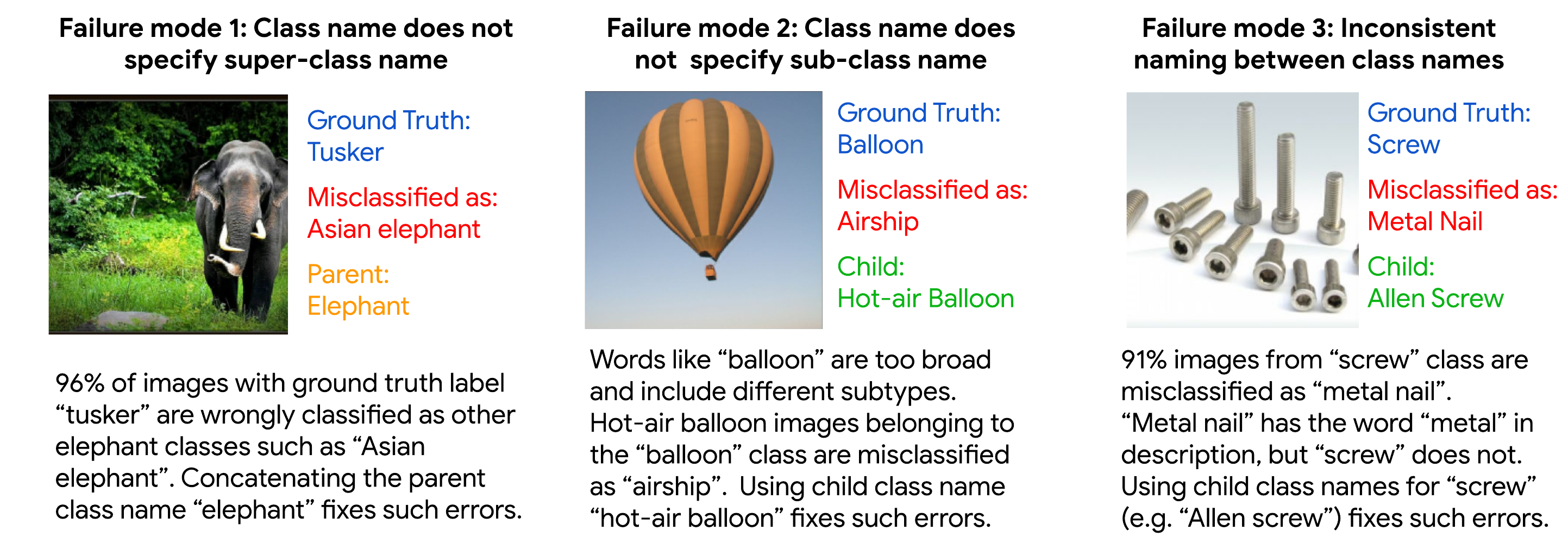}
\caption{Typical failure modes in the cases where top-5 prediction was correct but top-1 was wrong. 
}
\label{fig:failure-mode}
\vspace{-10pt}
\end{figure*}

We noticed a large gap between the top-$1$ accuracy and top-$5$ accuracy, 64.2\% vs. 89.4\% respectively, revealing potential headroom for improvement. 
We investigated the cases where the top-$1$ prediction was incorrect but the top-$5$ prediction was correct, and identified several typical failure modes. 
Despite the well-known multi-label issues in ImageNet \cite{beyer2020we}, we found many of the remaining failure cases are caused by noise and ambiguous text prompts related to the WordNet hierarchical structure of ImageNet.
Some class names are quite general so that the model cannot correctly match images from their specific subclasses. For example, the hot-air balloon images belonging to the ``balloon'' class were misclassified as ``airship'', see Figure \ref{fig:failure-mode} middle.
On the other hand, some class names are too specific such that the model fails to correlate them with their more generic super-classes. For example, 96\% of images with ground truth label ``tusker'' are wrongly classified as other elephant classes such as ``Asian elephant'', see Figure \ref{fig:failure-mode} left. 
The failure modes analysis suggests that the text encoder is very sensitive to inputs and as a result, the overall classification lacks robustness. 

Inspired by these observations, we propose to first identify the subset of images whose top-1 prediction is likely to be incorrect, and then improve the accuracy for those images by a principled framework to augment their class labels by WordNet hierarchy.
To estimate whether an image has an incorrect prediction, i.e., to estimate the prediction confidence, we use the consistency of predictions under different text prompt templates and image augmentations as a signal for prediction confidence estimation.
Although prediction confidence estimation has been well studied in single-modal classification models, we found those commonly used confidence scores, maximum softmax probability \citep{hendrycks2016baseline} and maximum logit score \citep{hendrycks2019scaling}, are not always reliable for the multi-modal CLIP and LiT models due to the poor calibration of the logits scores. 
For example, among the 1K classes in ImageNet, the class with the greatest mean logit value (computed as the cosine similarity between image and text embeddings) is ``fig'' (the fruit).
Though we don't have access to CLIP private training data, 
we hypothesize that this might be due to ``fig'' being a common abbreviation for ``figure'', which frequently occurs in the training data and thus includes many non-fruit illustrations.

In this work, we first propose a simple yet efficient zero-shot confidence estimation method better suited for CLIP, based on predictions' self-consistency over different text prompts and image perturbations. 
\citep{wang2022self} proposed using self-consistency among multiple model outputs to improve the reasoning accuracy of large language models. 
Here we extend the idea for confidence estimation in multi-modal models by measuring \emph{consistency of predictions under multiple input text prompts and image transformations}. Our method is effective at predicting mistakes;  
the identified low confidence subset has significantly lower top-1 accuracy (21.58\%) than the average accuracy (64.18\%). 
Next, to improve the accuracy for the low confidence subset, we develop a 
label augmentation technique using WordNet label hierarchy. Our method leverages semantic information from ancestors (top-down) as well as children (bottom-up) and improves the top-1 accuracy of the subset to 38.71\% (17.13\% improvement).
Our method not only improves model accuracy, but also model robustness, improving on ImageNet variants with distribution shift such as ImageNet-v2, ImageNet-R, ImageNet-Adversarial and Imagenet-Sketch. 

The main contributions of this work are:
\begin{compactitem}
    \item We identified several failure modes for zero-shot ImageNet classification using multi-modal models, and our findings suggest that the text encoder is very sensitive to prompts. To improve the prediction accuracy, prompts need to be better designed. 
    \item We propose a simple yet efficient zero-shot confidence score that is better suited for multi-modal models, based on predictions' self-consistency under different text prompts and image perturbations. 
    \item We develop a label augmentation technique that uses both ancestor and children labels from WordNet. By applying the label augmentation to the previously identified low confidence subset of images, we significantly improve their prediction accuracy. 
\end{compactitem}

\section{Related work}
%

\noindent \textbf{Confidence estimation.} Reliably estimating the confidence of a prediction is helpful for downstream decision making and can ensure the safe deployment of machine learning models. A well-calibrated confidence estimation should assign low scores for incorrect predictions and high score for correct predictions.
Maximum softmax probability \cite{hendrycks2016baseline} and maximum logit \cite{hendrycks2019scaling} are the most commonly used confidence scores for classification problems, because of their simplicity and computational efficiency. 
Recent works propose more sophisticated confidence estimation methods which either involve modifications to the classification models or significantly increase the inference time. For example,
Bayesian approaches such as Gaussian Process layer \cite{liu2022simple} and dropout-based variational inference \cite{gal2015dropout} assume the weights in the neural networks are random variables such that the final prediction follows a distribution. A large variance of a prediction indicates the low confidence of the prediction. 
Non-Bayesian methods such as ensemble-based methods which aggregate the predictions from multiple models to improve the robustness of the confidence estimation \cite{Lakshminarayanan2017, wen2020batchensemble}. 
Those sophisticated methods were developed and studied in the single-modal models, and the application to multi-modal models is not straightforward. In addition, those methods mostly require modification to the model and additional training, which becomes challenging to multi-modal models since the training data are generally not publicly available. 
In our work, we focus on a zero-shot confidence estimation that is exclusively designed for multi-modal models. 
Our method does not require additional training, and is simple, efficient, and effective. 

\noindent \textbf{Prompt engineering.}
Prompt engineering and learning has attracted much attention in vision and learning since the introduction of image-text models \cite{radford2021learning, jia2021scaling, zhai2022lit}. 
The image-text models align images and their text descriptions into a common space, which facilitates model generalization to unseen categories at inference time. 
However, it has been observed that downstream image classification accuracy highly depends on the specific input prompts. 
This motivates researchers to either fine-tune or auto-learn prompts when adapting multi-modal models to downstream vision tasks.
\cite{zhou2022learning, zhou2022conditional} propose CoOp and CoCoOp to automatically learn the prompt word embeddings in the few-shot settings, and show significant improvements over the vanilla zero-shot image classification based-on prompting. 
%
These are learning based approaches, requiring supervised data from downstream tasks, while our proposed method is zero-shot and post-hoc without using any supervised data. 
%
In concurrent work, \cite{shu2022test} proposes learning prompt embeddings in an unsupervised manner by minimizing the entropy of the averaged prediction probability distribution, where each prediction is based on a random augmentation applied to the input image.
Our work differs from \cite{shu2022test} in the sense that we do not learn an input-dependent prompt embedding. Instead we only selectively modify the prompts using knowledge hierarchy for images that have unreliable predictions, and our modified new prompt is natural language rather than a numerical embedding. 

\noindent \textbf{Label hierarchy.}
Label hierarchy or label ontology are relational graphs among semantic labels. WordNet is one of the most widely used concept ontologies, and it has been used for visual recognition problems. Fergus et al. \cite{fergus2010semantic} leverage the
WordNet hierarchy to define a semantic distance between any two categories and use this semantic distance to share labels. Deng et al. \cite{deng2014large} propose a hierarchy and exclusion graph to explicitly model the semantic relations among labels, and significantly improve object classification by exploiting the rich label hierarchy. The idea of semantic distance defined on the WordNet ontology graph is also used in \cite{rohrbach2011evaluating,rohrbach2010helps} for transferring knowledge in zero-shot learning problems. We are similar to the above work in that we utilize the label semantics encoded by the label hierarchy as well, but label hierarchy in our case is used in the multi-modality scenarios: textual labels and visual images are represented in the same latent space, therefore, the hierarchy structure is directly exploited in the representation space to steer the recognition process.  

\section{Zero-shot inference failure case analysis}
\label{sec:failure_analysis}



Given that the top-1  accuracy (64.2\%) is much lower than top-5 accuracy (89.4\%) for zero-shot ImageNet classification using CLIP, we investigated the failure cases that are ``top-5 correct but top-1 wrong" (12605 images, 25.2\% of all test images). 
Table. 1
in Suppl. shows some representative classes.
The failure modes are summarized as:

\noindent{\bf (1) Class name does not specify super-class name: }
Some classes, whose class names do not have their WordNet ancestor (e.g., ``tusker", one of 1k ImageNet classes, does not have its parent ``elephant" in the class name), may have a relatively lower score than other classes, which explicitly have the ancestor present in the class name (e.g., ``Asian elephant'').
See examples in Fig.~\ref{fig:failure-mode} (Left).\\ 
\noindent{\bf (2) Class name does not  specify sub-class name}: If the class name is too abstract, then its CLIP embedding is not necessarily close to the image embedding: e.g, CLIP wrongly classifies most images from ``balloon'' class as airship, see Fig.~\ref{fig:failure-mode} (Middle). That is because there are distinct kinds of balloons, each belonging to a different semantic subgroup. Relying on the text embedding of the fine-grained children's class names (e.g., using ``hot-air balloon'') often fixes these errors. \cite{beyer2020we} reported the similar issue of label ambiguity in ImageNet. \\
\noindent{\bf (3) Inconsistent naming between class names: 
} Some ImageNet class names are nouns, but  others are adjective-prefixed nouns. This may make CLIP text embedding biased, see one example in Fig.~\ref{fig:failure-mode} (Right) where images from ``screw'' class are misclassified as ``metal nail''.

\begin{figure*}[t]
\vspace{-15pt}
\centering
 \includegraphics[width=\textwidth]{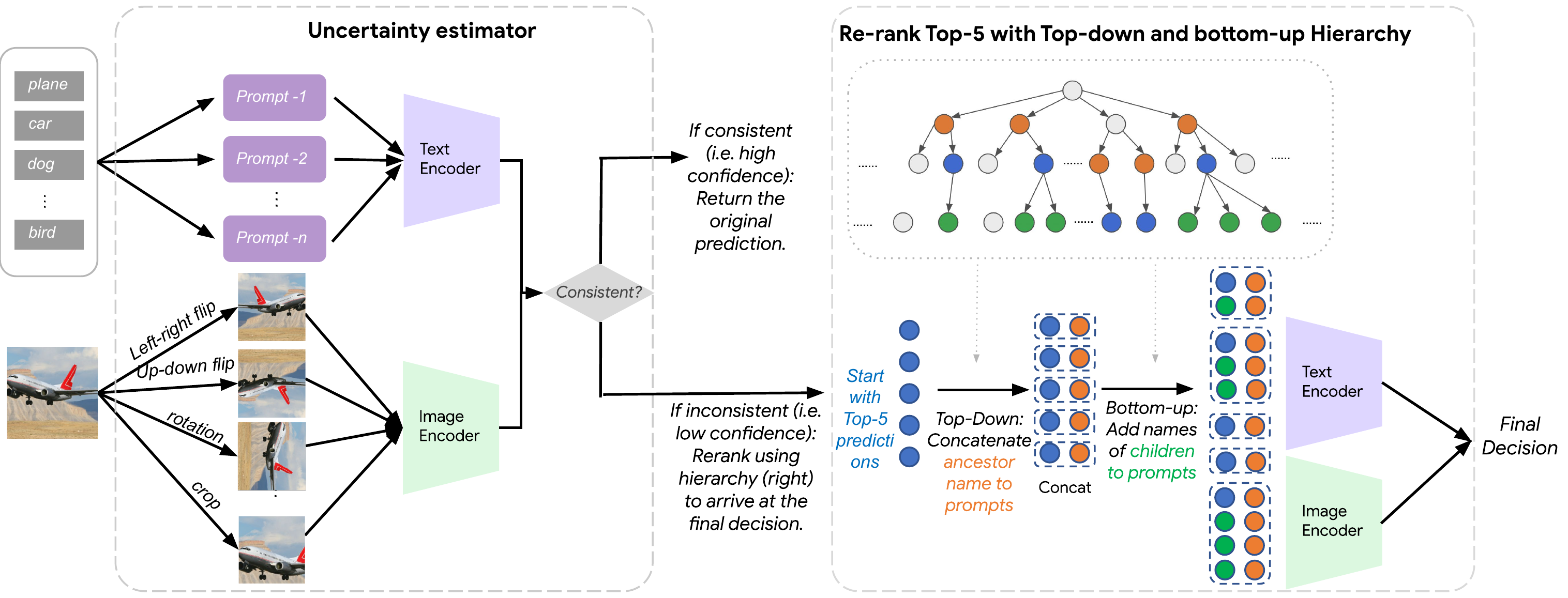}
\caption{ Our zero-shot classification pipeline consists of 2 steps: confidence estimation via self-consistency (left block) and top-down and bottom-up label augmentation using the WordNet hierarchy (right block). See
Algorithms~\ref{alg:uncertainty} and \ref{alg:hierarchy} 
for pseudocode.
}
\label{fig:method}
\vspace{-10pt}
\end{figure*}

\section{Proposed Method}

As shown in Section \ref{sec:failure_analysis}, CLIP models can be sensitive to different text prompts for images in certain classes. 
In this section, we first propose a confidence estimation method to identify low confidence predictions. 
We show that the identified subset has much lower accuracy than the average (Sec.\ref{sec:confidence_estimation}).  
We next develop a principled method that utilizes knowledge hierarchy to improve the accuracy of the low confidence subset, and consequently improve the overall accuracy on the whole datasets (Sec.~\ref{sec:method_hierachy}).  
\subsection{Self-consistent zero-shot confidence estimation}
\label{sec:confidence_estimation}
Given an image $\vx$ and a candidate class name $c$, where $c\in \mathcal{C}, |\mathcal{C}|=1000$, the CLIP model encodes $\vx$ and $c$  respectively by its image encoder $f_{image}$ and text encoder $f_{text}$, denoted as $\vz_m=f_{image}(\vx)$ and $\vz_c=f_{text}(c)$. 
The prediction logit score is  defined as $\text{logit}(\vx, c) = \cos(\vz_m, \vz_c)$,  where $\cos(\cdot, \cdot)$ is the cosine similarity between two vectors, and the predicted class is $\argmax_{c\in\mathcal{C}} \text{logit}(\vx, c)$.
We estimate the confidence by the self-consistency rate when applying different context prompts and image augmentations. 

\vspace{-10pt}
\paragraph{Confidence estimation via text prompts.}
To improve the zero-shot classifier's performance, the CLIP paper \citep{radford2021learning} hand crafted various context prompts, e.g. ``\texttt{A photo of a big \{label\}}'' and ``\texttt{A photo of a small \{label}\}''), for different datasets for the purpose of prompt ensembling: 
For an image $\vx$, given a set of context prompts $\mathcal{T}$, the ensembled logit score is $\text{logit}(\vx, \mathcal{T}(c))=\frac{1}{|\mathcal{T}|}\sum_{t\in\mathcal{T}} \text{logit}(\vx, t(c))$, where $t(c)$ denotes the new prompt after applying context prompt $t(\cdot)$ to $c$. 
Here instead of using the prompts for ensembling, we make use of the prompts to define our confidence score. 
Given a set of prompts $\mathcal{T}$, we apply each of the prompt $t(\cdot)$ for the classifier, and see if the top-1 prediction is the same as that when applying no prompt. 
We use the percentage of prompts that have consistent top-1 prediction with that without prompt as the confidence score $S_{\mathcal{T}}(\vx)$, i.e.
\begin{align}
    S_{\mathcal{T}}(\vx) = \frac{ \sum_{t\in\mathcal{T}} \mathbbm{1}\{ \hat{c}(\vx, t) = \hat{c}(\vx, \emptyset) \} }{|\mathcal{T}|}
    \label{eq:conf_prompt}
\end{align}
where $\hat{c}(\vx, \emptyset) = \argmax_{c \in \mathcal{C}}  \text{logit}(\vx, c)$ is the top-1 prediction using the pure class name, and $\hat{c}(\vx, t) = \argmax_{c \in \mathcal{C}} \text{logit}(\vx, t(c))$ is the top-1 prediction when applying prompt $t(\cdot)$.
Intuitively, a reliable prediction should have highly consistent top-1 predictions when context prompts are applied or not, and therefore should have a high confidence score $S_{\mathcal{T}}(\vx)$ with respect to the prompt set $\mathcal{T}$, and vice versa. 

\vspace{-10pt}
\paragraph{Confidence estimation via image perturbation.} 
We can also estimate the confidence of a prediction based on the self-consistency when applying different perturbations to the input image.
Intuitively, if the top-1 predictions are inconsistent when applying different image perturbations, the prediction is unreliable. 
Specifically, we consider the common image transformations, left-right flip, rotation, crop, etc., and apply the perturbation method $b(\cdot)$ to the input image, and infer the predicted class as $\hat{c}(\vx, b) = \argmax_{c\in \mathcal{C}}\text{logit}(b(\vx), c)$. 
We define the confidence score with respect to a set of image perturbations $\mathcal{B}$ as, 
\begin{align}
    S_{\mathcal{B}}(\vx) = \frac{ \sum_{b\in\mathcal{B}} \mathbbm{1}\{ \hat{c}(\vx, b) = \hat{c}(\vx, \emptyset) \} }{|\mathcal{B}|}
    \label{eq:conf_image}
\end{align}
We expect a high confidence prediction to have highly consistent prediction when applying different image perturbations, and therefore to have a high confidence score $S_{\mathcal{B}}(\vx)$ with respect to the image perturbation set $B$. 

\vspace{-10pt}
\paragraph{Determining the low confidence subset by combining the two confidence estimations.}
The confidence score we proposed in Eq. (\ref{eq:conf_prompt}) and Eq. (\ref{eq:conf_image}) are continuous values. A threshold needs to be determined if we want to select a subset of examples with low confidence using the continuous confidence score. 
In practice, the threshold can be chosen based on the requirement of recall and precision trade-off in the real application.
In our study, to bypass the threshold selection, we propose to use a binary criterion for determining the low confidence set. 

For IamgeNet dataset, the CLIP paper \cite{radford2021learning} designed total 80 context prompts. 
We define four sets based on the 80 prompts: the first 40 prompts $\mathcal{T}_1$, the last 40 prompts $\mathcal{T}_2$, all 80 prompts $\mathcal{T}_3$, and no prompts $\mathcal{T}_4=\emptyset$. 
We apply the four different sets of prompts to the classifier and see if their top-1 predictions are all consistent or not, i.e. $\hat{c}(\vx, \mathcal{T}_1) = \hat{c}(\vx, \mathcal{T}_2) = \hat{c}(\vx, \mathcal{T}_3) = \hat{c}(\vx, \mathcal{T}_4)$. 
Then we determine the low confidence subset $\mathcal{O}_\mathcal{T}$ as those examples who have inconsistent predictions among the 4 prompts sets.
We studied other choices such as using a random set of 40 prompts as $\mathcal{T}_1$, or splitting the 80 prompts into more subgroups, and found the results were very similar. 

Similarly we also determine a low confidence subset $\mathcal{O}_\mathcal{B}$ based on image perturbations. 
In practice we found left-right flip works the best among the above mentioned perturbations. 
Thus for simplicity, we compare the top-1 prediction when applying the left-right flip to the input image and the top-1 prediction when using raw image. If their predictions are not consistent, that example will be included into the low confidence set $\mathcal{O}_\mathcal{B}$. 

Finally, we use the union of the two low confidence sets $\mathcal{O}_\mathcal{T}$ identified using the text prompts and and $\mathcal{O}_\mathcal{B}$ identified using the image perturbations as the final low confidence subset $\mathcal{O}$ in the following
experiments. 
Algorithm \ref{alg:uncertainty} shows the low confidence set generation process.

\begin{algorithm}[tp]
\small
\caption{Zero-shot confidence estimation}
\label{alg:uncertainty}
\SetKwInput{KwData}{Input}
\SetKwInput{KwResult}{Output}

\KwData{Input images $ \mathcal{X}=\{\vx_i\}_{i=1}^N$, Candidate class set $\mathcal{C}$, image encoder $f_{image}$ and text encoder $f_{text}$, text threshold $\tau_{t}$, image threshold $\tau_{i}$}
\KwResult{Low confidence set $\mathcal{O}$}

\SetAlgoLined
\nl
Low confidence set $\mathcal{O}_{\mathcal{T}} \gets \emptyset$ \ \algcomment{Confidence estimation via text prompts}

\nl
Sample $L$ different context prompt $t_1$, $t_2$ \ldots  $t_L$


\nl
\For{$\vx_i \in \mathcal{X}$}{
\nl
Compute $S_{\mathcal{T}}(\vx_i)$ based on Eq. (\ref{eq:conf_prompt})


\nl
\If { $S_{\mathcal{T}}(\vx_i)$ $>$ $\tau_{t}$}{
$\vx_i$ has  high confidence prediction
    }
\Else {$\mathcal{O}_{\mathcal{T}} \gets \mathcal{O}_{\mathcal{T}} \cup \vx_i$
}
}

\nl
Low confidence set $\mathcal{O}_{\mathcal{B}} \gets \emptyset$ \ \  \algcomment{Confidence estimation via  image perturbation}

\nl
Sample $M$ perturbation methods $b_1, \dots, b_M$ 

\nl
\For{$\vx_i \in \mathcal{X}$}{

Compute $S_{\mathcal{B}}(\vx_i)$ based on Eq. (\ref{eq:conf_image})

\nl
\If {$S_{\mathcal{B}}(\vx_i)$ $>$ $\tau_{i}$}{
$\vx_i$ has high confidence prediction
    }
\Else {$\mathcal{O}_{\mathcal{B}} \gets \mathcal{O}_{\mathcal{B}} \cup \vx_i$
}
}

\nl
$\mathcal{O} \gets \mathcal{O}_{\mathcal{T}} \cup \mathcal{O}_{\mathcal{B}}$
\end{algorithm}

\subsection{Top-down and bottom-up label augmentation using WordNet hierarchy}
\label{sec:method_hierachy}
Through extensive 
analysis of the incorrect predictions among the identified unreliable predictions, we 
found that many of them are caused by CLIP's lack of robustness to prompts. Instead of tuning the prompt templates, we focus on how to augment {\small\texttt{\{label\}}} in ``{\small\texttt{A photo of a \{label\}}}''. 
A proper prompt that specifies both the generic type and the more specific sub-types of this class are very important for correctly classifying the image.
However, the ImageNet \cite{deng2009imagenet} class names are not all defined with similar specificity and some classes are more abstract than others, e.g. 350 classes have children, while the rest of the classes have no children. See Suppl. Fig. 1 
for more details. 
To make the ImageNet classification problem better suited to CLIP, we leverage the underlying WordNet hierarchy and develop a top-down and bottom-up class name augmentation method to improve zero-shot prediction accuracy for unreliable predictions.

The WordNet hierarchy is a semantic concept ontology, with nodes being cognitive synonyms indicating different concepts, and edges indicating the super-subordinate relation between concepts. Traveling upward from leaf nodes to the root, the concepts start from the very specific to the generic. 
For example, starting from the edge node ``strawberry'' to the root are ``berry'',  ``edible fruit'', ``produce'', ``food'', ``solid'', ``matter'', and ``physical entity'' (the root).
As we have seen in the failure mode analysis, many of the imageNet class names suffer from either being too abstract or being too specific, so that their concepts do not align well with the visual concepts the CLIP model learned in training. We propose using the WordNet knowledge hierarchy to augment the class labels in prompts so that the CLIP model has a better match between the image and prompts. 

\vspace{-10pt}
\paragraph{Top-down: augmenting class names with parent.}
As shown in failure case analysis, adding the super-class name to reduce ambiguity and to encourage the model's attention on the generic concept is helpful for improving the accuracy. 
Therefore we propose using WordNet to find the parent node of the raw class name, and concatenate it to the class name, i.e.
$
    \text{logit}(\vx, c) = \text{logit}(\vx, [c; p(c)])
$
where $p(c)$ is the parent node's name of the class name $c$, and $[c; p(c)]$ means the string concatenation of the class name and the parent name.
We apply the method to top-5 predicted classes. Using the newly defined class names, we are able to re-rank the top-5 predictions for the identified unreliable subset of images. 
Note that WordNet contains a few very abstract class names for nodes, such as ``physical entity'', ``artifact'', ``matter'', etc. 
We found that such parent nodes are not informative, hence we remove them.
There are also many academic words in WordNet, for example the parent node of sea anemone is ``anthozoan'',
 which can be rare in CLIP training data. 
Adding those academic words to class name makes the prediction even less robust. 
So we simplify the WordNet by pruning based on an estimation of the word frequency in CLIP training data by using embedding norm. 


\vspace{-10pt}
\paragraph{Bottom-up: augmenting class names with children.}
Some ImageNet class names are generally abstract, but the ImageNet images may belong to a specific subtype of the class. For example, ``balloon'' is a class name in ImageNet, but most balloon images in ImageNet are actually ``hot-air balloon'', which is a child of ``balloon'' in WordNet hierarchy. The logit score for a parent class is not necessarily higher than the score for its child classes, mismatching with hierarchy prior. To accurately classify the images using CLIP, we need to augment the class name with fine-grained child subclasses. 
For each class $c$ having children in the WordNet hierarchy, we redefine the logit score as the max score over itself and all its children, i.e.,
$
    \text{logit}(\vx, c) = \max \{\text{logit}(\vx, c), \text{logit}(\vx, c_1), \dots, \text{logit}(\vx, c_r)\},
$
where $c_1 \dots c_r$ are the $r$ children of the node $c$ in the WordNet hierarchy. 
We apply this bottom-up method to top-$5$ predicted class names, and re-rank the top predictions. 

\vspace{-10pt}
\paragraph{Combining Top-down and bottom-up.}
In practice, we use both children and the ancestor(parent) to augment each class $c$, to transfer semantic information bidirectionally in both top-down and bottom-up way: the ancestor(parent) class is more generic than $c$, and has better chance to disambiguate instance from a more abstract level; on the other hand, children categories have more specific attribute description, and the attribute descriptions are semantically meaningful representations bridging the gap between the image embedding and its abstract class concept $c$. Then the final logit score between $x$ and $c$ is: 
\begin{multline}
    \text{logit}(\vx, c) =\max \{\text{logit}(\vx, [c; p(c)]), \\ \text{logit}(\vx, [c_1; p(c)]), \dots, \text{logit}(\vx, [c_r; p(c)])\}
    \label{eq:logit_tp_bu}
\end{multline}
where $p(c)$ is parent of $c$, and $c_1 \dots c_r$ are $c$'s children. The $\hat{c}$, where $\hat{c}\in \mathcal{C}_{top5}$, with the maximal logit score is the predicted class of $\vx$. 
See Algorithm \ref{alg:hierarchy} for details.

\begin{algorithm}[tp]
\small
\caption{Top-down and bottom-up class label augmentation using WordNet hierarchy}
\label{alg:hierarchy}
\SetKwInput{KwData}{Input}
\SetKwInput{KwResult}{Output}

\KwData{Input image $\vx \in \mathcal{O}$, top-5 candidate class set $\mathcal{C}_{top5}$, sparse WordNet hierarchy $H$, image encoder $f_{image}$ and text encoder $f_{text}$}
\KwResult{Predicted class of $\vx$}

\SetAlgoLined

\nl
Candidate class set $\mathcal{C} \gets \emptyset$ 

\nl
\For{$c \in \mathcal{C}_{top5}$}{

$\mathcal{C} \gets \mathcal{C} \cup [c; \textnormal{parent}(c)]$, where $\textnormal{parent}(c)$ is the parent of $c$ in $H$
\ \  \algcomment{Top-down}

\nl
\If{$c$ \textnormal{has} $r \geq 1$ \textnormal{children} $c_1 \dots c_r$ in H }{
$\mathcal{C} \gets \mathcal{C} \cup \{[c_j; \textnormal{parent}(c)]\}^{r}_{j=1}$ \ \  \algcomment{Bottom-up}
}
}

\nl
$\hat{c} \gets \argmax_{c \in \mathcal{C}} \textnormal{logit}(\vx, c)$

\If{$\hat{c}\in \mathcal{C}_{top5}$}{ final prediction $\gets \hat{c}$ }
\Else {final prediction $\gets \textnormal{parent}(\hat{c})$}
\end{algorithm}

\section{Experiments and Results}
\begin{figure*}[htb]
\centering
\begin{subfigure}[t]{0.35\textwidth} 
\centering
\caption{CLIP: Calibration ROC and AUC}
\vspace{-1.5em}
\includegraphics[width=0.96\textwidth]{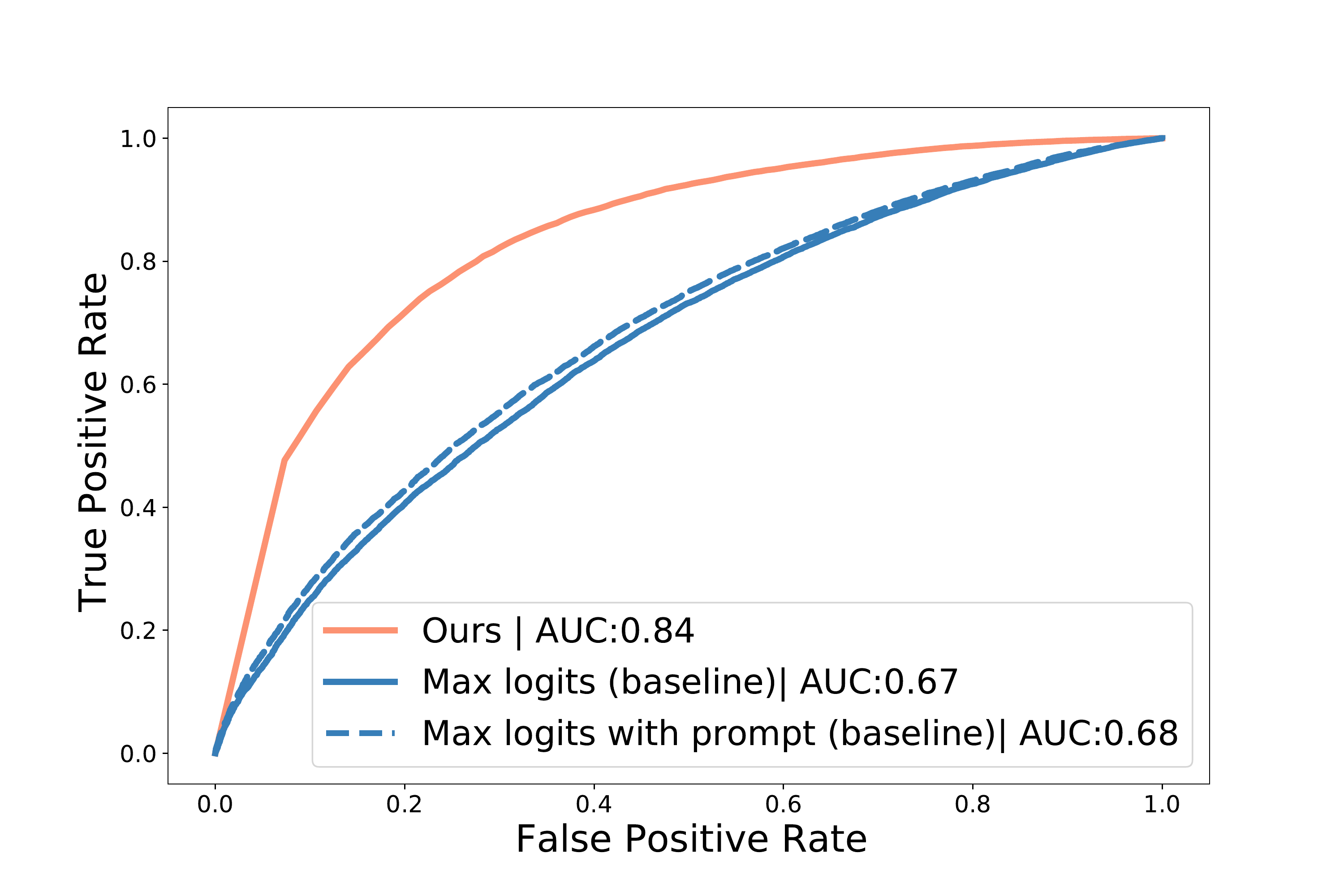}
\label{fig:clip_auc}
\end{subfigure} 
\begin{subfigure}[t]{0.35\textwidth} 
\centering
\caption{CLIP: Selective prediction}
\includegraphics[width=0.84\textwidth]{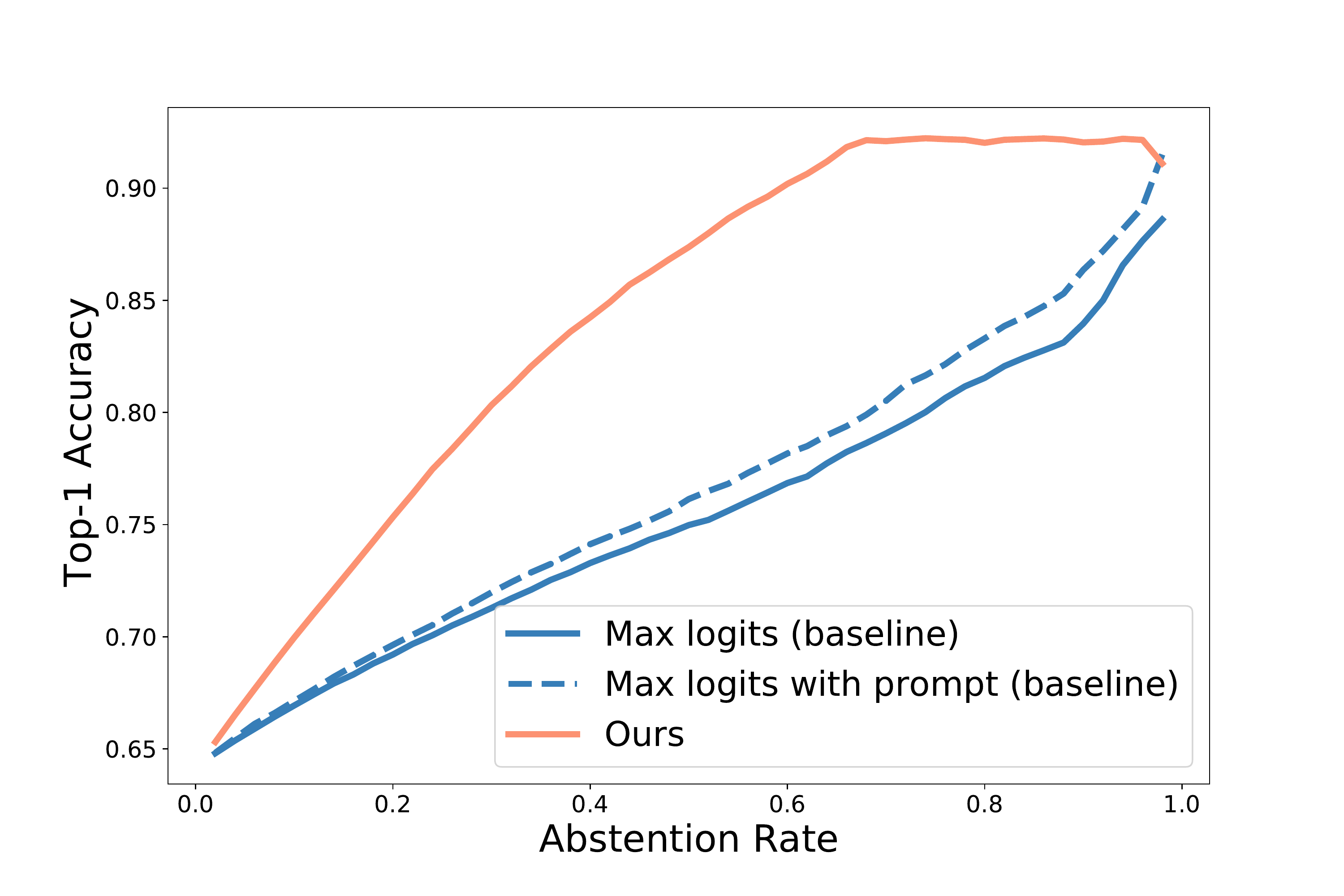} 
\label{fig:clip_sp}
\end{subfigure} 
\begin{subfigure}[t]{0.35\textwidth} 
\centering
\caption{LiT: Calibration ROC and AUC}
\includegraphics[width=0.84\textwidth]{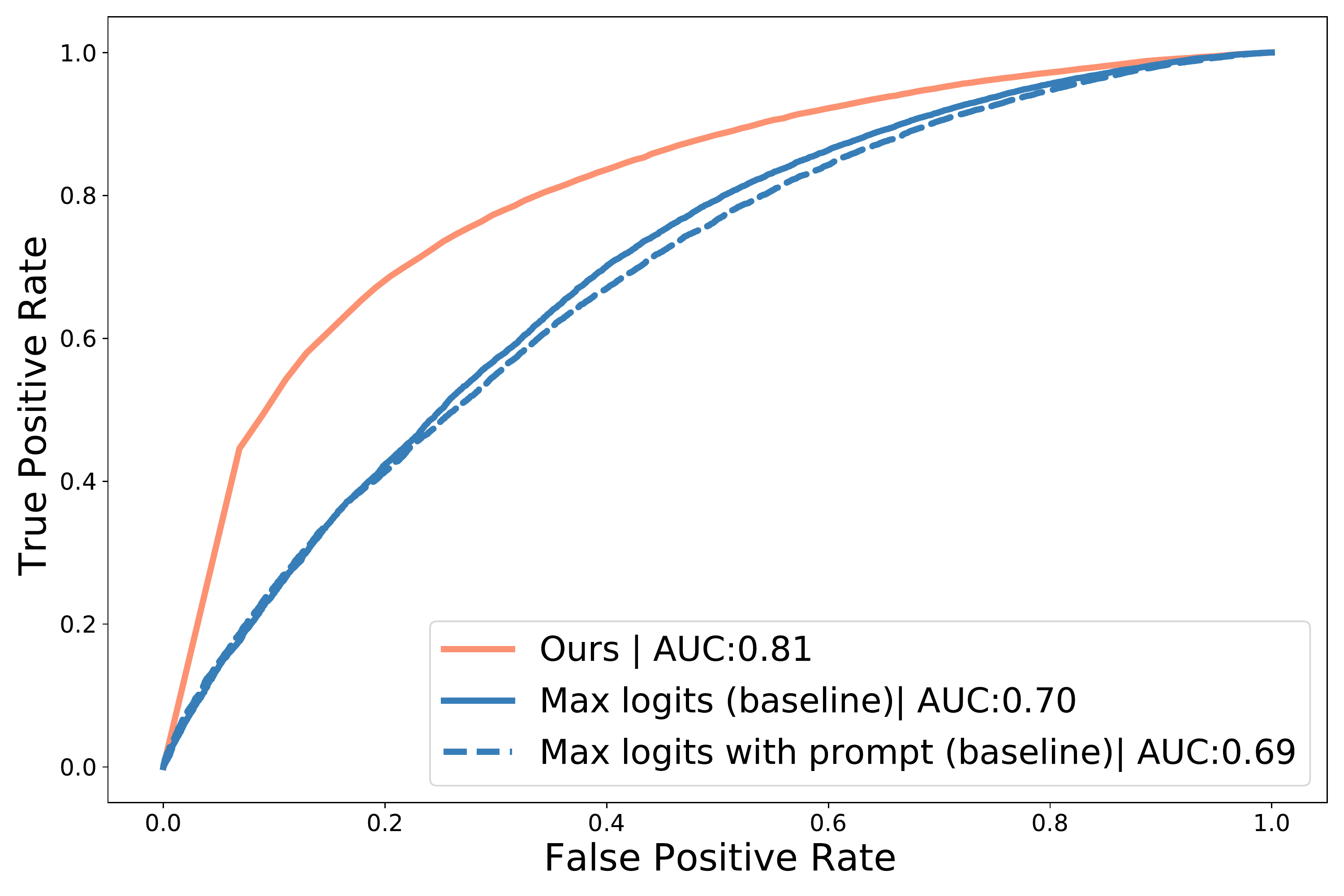}
\label{fig:lit_auc}
\end{subfigure} 
\begin{subfigure}[t]{0.35\textwidth} 
\centering
\caption{LiT: Selective Prediction}
\includegraphics[width=0.84\textwidth]{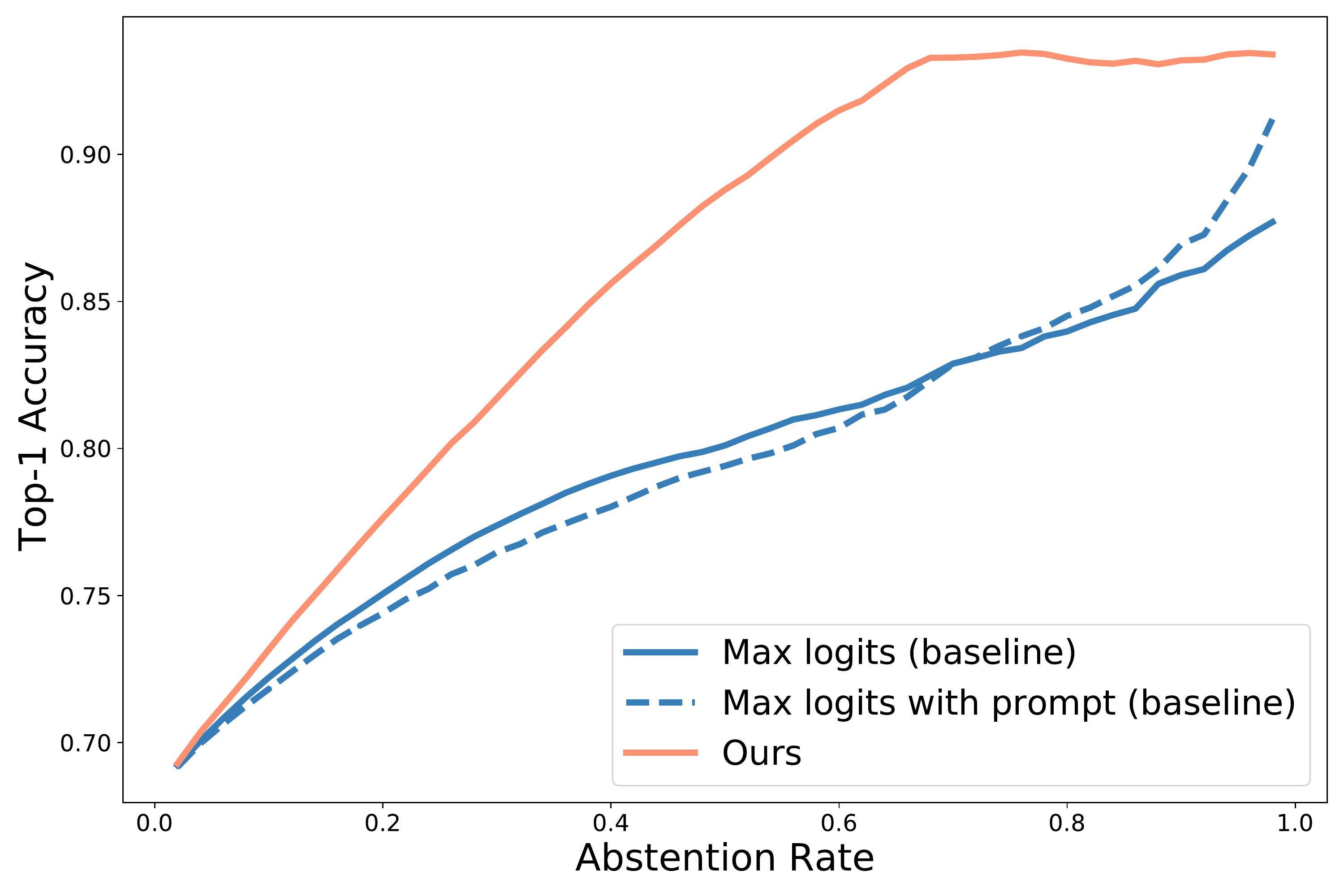} 
\label{fig:lit_sp}
\end{subfigure} 
\caption{ROC plots (left column) show that our proposed confidence score is better at distinguishing correct and incorrect predictions and results in higher AUC scores than baselines for both CLIP (ViT-B/16) (a) and LiT (ViT-B/32)(c). Selective prediction curves (right column) show that our proposed confidence score is better at abstaining incorrect predictions and as a result the accuracy of the remaining set is higher than the baselines for both CLIP (ViT-B/16) (b) and LiT (ViT-B/32) (d). }
\label{fig:conf_esti}
\end{figure*}

Our proposed method is composed of two steps and we conduct experiments to verify the effectiveness of each step: 
(1) Use zero-shot confidence estimation to identify the low confidence subset of samples (see Fig.~\ref{fig:conf_esti} for the results), and
(2) Augment the class label using top-down and bottom-up strategies based on the sparsified WordNet on the low confidence subset to improve the accuracy (See Table~\ref{table:zero-shot} and Table~\ref{table:otherdataset} for the results).

\begin{table*}[tp]
\large
\centering
\caption{CLIP (ViT-B/16) and LiT (ViT-B/32) zero-shot top-1 accuracy comparison between baseline and ours (w/ hierarchy).}
\scalebox{.7}{%
\begin{tabular}{l c |c c| c c}
    \toprule
     & & CLIP & (Ours) Hierarchy-CLIP & LiT & (Ours) Hierarchy-LiT\\
    \midrule
     \multirow{2}{*}{ImageNet \cite{deng2009imagenet}} & Low conf. set  & 21.58\% & \bf{38.71\%}  & 31.18\% & \bf{37.25\%} \\ 
     & Full set & 64.18\% & \bf{67.78\% } & 68.26\% & \bf{69.41\% }  \\ 

    \midrule
     \multirow{2}{*}{ImageNet-v2 \cite{recht2019imagenet}}& Low conf. set  & 17.77\% & \bf{32.50\%} & 27.08\% & \bf{31.45\% }\\
     & Full set  & 58.06\% & \bf{61.07\%} & 60.11\% & \bf{61.11\% } \\
    
    \midrule
     \multirow{2}{*}{ImageNet-R \cite{hendrycks2021many}} & Low conf. set  & 16.79\% & \bf{27.91\% } & 21.82\% & \bf{22.93\% } \\
     & Full set & 56.88\% & \bf{59.46\%} & 66.54\% & \bf{66.75\% }  \\
    
    \midrule
     \multirow{2}{*}{ImageNet-Adversarial \cite{hendrycks2021natural}} & Low conf. set  & 10.13\% & \bf{18.44\% } & 7.19\% & \bf{8.95\% }\\
     & Full set & 26.12\% & \bf{29.23\% } & 13.93\% & \bf{14.56\%}  \\
     
    \midrule
     \multirow{2}{*}{ImageNet-Sketch \cite{wang2019learning}} & Low conf set & 13.74\% & \bf{23.18\% } & 21.51\% & \bf{24.42\%}\\
     & Full set  & 44.71\% & \bf{47.28\% } & 52.47\% & \bf{53.17\% }  \\   
     
     \bottomrule
\end{tabular}
}
\label{table:zero-shot}
\end{table*}

\subsection{Our proposed confidence score is better suited for selective prediction than baselines}

A well-calibrated confidence estimator should score high for those correct predictions, and low for incorrect predictions. 
As a result, a good confidence estimator should be a good predictor for prediction correctness. We plot the receiver operating characteristic (ROC) curve and compute the area under the ROC curve (AUC) as a quantitative measure to compare our proposed confidence estimation with the baselines. 
An AUROC of 1.0 indicates perfect separation between correct and incorrect predictions, and 0.5 means the two groups are not distinguishable.
Maximum logit score, $\max_{c\in\mathcal{C}} \text{logit}(\vx, c)$ is one of the most commonly used confidence score for classification problems in single modal models \cite{hendrycks2019scaling}, so we consider it as our baseline. 
Fig. \ref{fig:clip_auc} and \ref{fig:lit_auc} clearly show that our confidence score is significantly better than the baseline method at distinguishing between correct and incorrect predictions, for both CLIP and LiT models. 
The AUC score for our proposed method is above 0.8 while that for the baseline method is around 0.7.

We also compare our method with the baseline in the scenario of selective prediction.
Given a budget of abstention rate $\alpha\%$, the best strategy is to abstain the $\alpha\%$ samples with the lowest confidence scores. If the confidence score is well calibrated, the accuracy for the abstained set will be low and as an evidence the accuracy of the remaining set would be high.
We plot the selective prediction curves \citep{Lakshminarayanan2017}, which reports the accuracy on the remaining set as a function of the abstention rate. 
Fig. \ref{fig:clip_sp} and \ref{fig:lit_sp} show that our proposed confidence score results in higher accuracy than the baseline maximum logit score at all abstention rates for both CLIP and LiT.

Prompt ensemble has been shown to improve accuracy and robustness of the prediction, so here we also compare ours with the maximum logit score after applying prompt ensemble. As shown in the selective prediction curves, although the prompt ensemble indeed helps to achieve higher accuracy (dashed line) than that using the pure class name (solid line), it is still inferior to our proposed method.

\vspace{-3pt}
\subsection{Using hierarchy to help improve zero-shot accuracy on low confidence subset}
\vspace{-3pt}
Using top-down and bottom-up label augmentation significantly improves the accuracy on the low confidence subset.
We apply the top-down and bottom-up label augmentation on the low confidence subset: to better combine child and parent name, we create a prompt template to transform the child and parent name pairs into a new class name $\tilde{c}$ in natural language: ``{\small\texttt{\{child\} which is a kind of \{parent\}}}'' (different prompt templates may have different results).
Table~\ref{table:zero-shot} shows improvement of 17.13\% on the top-1 accuracy (from 21.58\% to 38.71\%) for the identified low confidence subset of samples, and overall 3.6\% on the top-1 accuracy (64.18\% to 67.78\%) for all samples in ImageNet.
We show similar improvement on the zero-shot accuracy for ImageNet shifted datasets. 
To investigate if our method works for other multi-modal models, we apply it to the LiT \citep{zhai2022lit} model and observe that our method improves accuracy for LiT models as well. See Supp. Fig. 2 
for qualitative visualization. 

\vspace{-10pt}
\paragraph{Generalizability to non-ImageNet datasets}
To show the generalizability of our methods on non-ImageNet datasets, 
We conducted experiments on 4 additional datasets: Caltech-101 \cite{li2003caltech101} (101 categories), Flower-102 \cite{nilsback2008automated} (102 flower categories), Food-101 \cite{bossard14} (101 food categories) and Cifar-100 \cite{krizhevsky2009learning} (100 categories). For each dataset, a subset of their categories are exist/aligned with WordNet hierarchy, we only apply our method on those WordNet aligned class names, where we could find their ancestor and children. We keep the other class names unmodified. We use CLIP (ViT-B/16) as multi-modal model. Table~\ref{table:otherdataset} shows that our method consistently improved accuracy on the low-confidence set (low) and the entire set (full):

\begin{table}[h]
\begin{center}
\footnotesize
\vspace{-10pt}
\caption{Generalizability to non-ImageNet datasets (CLIP (ViT-B/16) zero-shot top-1 accuracy). }
\resizebox{0.48\textwidth}{!}{
\begin{tabular}{c|c|c|c|c}
\hline
Dataset & orig (low) & ours (low) & orig (full) & ours (full) \\

\hline
Caltech-101 \cite{li2003caltech101} & 10.6 \% & 27.2\% (+16.6\%) & 74.1\% & 77.1\% (+3.0\%)\\
\hline
Flower102 \cite{nilsback2008automated}& 20.0\% & 29.4\% (+9.4\%) & 63.7\% & 65.3\% (+1.6\%)\\
\hline
Food-101 \cite{bossard14}& 28.2\% & 49.0\% (+20.8\%) & 84.7\% & 86.8\% (+2.1\%)\\
\hline
Cifar-100 \cite{krizhevsky2009learning}& 9.4\% & 17.5\% (+8.1\%) & 31.8\% & 35.2\% (+3.4\%) \\
\hline
\end{tabular}
}
\label{table:otherdataset}
\end{center}
\vspace{-15pt}
\end{table}

\subsection{Ablation study}
\paragraph{Generalizability to other backbones}
\noindent To study the generalization of our method to different model architectures and sizes, we used 4 additional backbones of CLIP, including convolutional neural network (CNN) based backbones (ResNet-50, ResNet-101) and vision transformer (ViT) based backbones (ViT-B/32, ViT-B/16 and ViT-l/14). Table~\ref{table:backbone} shows the improved accuracy after using our method on ImageNet with CLIP models of different backbones. Our method achieves consistently improved accuracies.

\begin{table}[t]
\begin{center}
\vspace{-10pt}
\caption{Generalizability to different backbones with CLIP. }
\small
\resizebox{0.48\textwidth}{!}{
\begin{tabular}{c|c|c|c|c|c}
\hline
backbone & ResNet-50 & ResNet-101 & ViT-B/32 & ViT-B/16 & ViT-l/14\\
\hline
ACC (low) & +14.25\% & +12.97\% & +15.12\% & + 17.13\% &+18.89\%\\
\hline
ACC (full) & +3.73\% & +3.71\% & +3.65\% & + 3.60\% & +3.23\%\\
\hline
\end{tabular}
\label{table:backbone}
}
\end{center}
\vspace{-10pt}
\end{table}

\begin{table}[t]
\centering
\vspace{-5pt}
\caption{CLIP (ViT-B-16) zero-shot top-1 accuracy comparison with prompt ensemble.}
\scalebox{0.65}{%
\begin{tabular}{l c |c c}
    \toprule
     & & Ensemble only & Hierarchy and Ensemble \\
    \midrule
     \multirow{2}{*}{ImageNet \cite{recht2019imagenet}} & Low conf. set  & 41.05\% & \bf{42.09\%}  \\ 
     & Full set & 68.48\% & \bf{68.86\% } \\ 

    \midrule
     \multirow{2}{*}{ImageNet-v2 \cite{recht2019imagenet}}& Low conf. set  & \bf{36.39}\% &  36.34\% \\
     & Full set  &   \bf{62.02\%} & 62.00\% \\
    
    \midrule
     \multirow{2}{*}{ImageNet-R \cite{hendrycks2021many}} & Low conf. set  &  35.13\% & \bf{ 36.12\% }  \\
     & Full set &  60.21\% & \bf{ 60.62\%}   \\
    
    \midrule
     \multirow{2}{*}{ImageNet-Adversarial \cite{hendrycks2021natural}} & Low conf. set  &  21.13\% & \bf{ 22.00\% } \\
     & Full set &  30.59\% & \bf{ 31.07\% }   \\
     
    \midrule
     \multirow{2}{*}{ImageNet-Sketch \cite{wang2019learning}} & Low conf. set &  \bf{27.13}\% &  26.56\% \\
     & Full set  &  \bf{48.52\%} & 48.26\%   \\   
     
     \bottomrule
\end{tabular}
}
\label{table:add_prompt}
\vspace{-12pt}
\end{table}

\vspace{-10pt}
\paragraph{Our hierarchy-based label augmentation is complimentary to prompt ensembling.} 
Prompt ensembling (PE) \cite{radford2021learning} requires a set of manually crafted prompt templates, and the zero-shot performance is sensitive to the set of prompts the model uses. 
Alternatively, our proposed method does not require a dedicated tuning of the prompt templates. We directly augment the class name with knowledge of the hierarchy from WordNet.  In addition, PE is computationally intensive because it needs to infer the embeddings of 80 prompt templates where each is applied with 1000 ImageNet classes, while our method only need to infer once for each of the predicted top-5 labels. Our method is more straightforward and interpretable given that it clearly shows the contribution of parent/child in the decision. 
Intuitively, PE is typically focused on fixing \textit{\{class\}} and augmenting contextual templates, while our method augments the \textit{\{class\}} with a fixed contextual template.
To verify if our hierarchy-based method is complimentary with prompt ensembling, we apply prompt ensembling after applying our top-down and bottom-up label augmentation. 
For the low confidence set, we first create a prompt template to transform the child and parent name pairs into a new class name $\tilde{c}$ in natural language: ``{\small\texttt{\{child\} which is a kind of \{parent\}}}''. Then we apply the 80 prompts designed by the CLIP paper \cite{radford2021learning} individually to the new class name $\tilde{c}$, and then ensemble them. 
For the high confidence set, since we do not modify the class name using hierarchy information, we only apply the prompt ensemble. 
The performance is shown in Table \ref{table:add_prompt}. We compare the zero-shot accuracy using the vanilla prompt ensembling method proposed in CLIP, and the zero-shot accuracy using our combined version of hierarchy-based class name augmentation and prompt ensembling. 
As shown in the table, using both hierarchy and prompt ensembling achieves better or on par accuracy with the prompt ensemble alone, suggesting that the two methods can be combined. 
Considering the prompt ensemble requires manually designed prompt templates and much greater inference time, our hierarchy-based class name augmentation is simple, efficient and effective. 
We also computed IoU of corrected low-confidence instances (\textit{low set}) between PE and our method: the IoU is 0.55, which implies the two methods are complementary for fixing errors. 

\vspace{-10pt}
\paragraph{Effect of threshold of confidence score on zero-shot accuracy.}
In Table \ref{table:zero-shot} we use a binary criterion to determine the low confidence set. 
We can alternatively use the continuous confidence score by choosing a threshold based on the trade-off between precision and recall. 
Changing the threshold of the confidence score can lead to different numbers of samples in the low confidence set.
We study the effect of threshold on zero-shot accuracy. Table \ref{table:threshold} shows the overall accuracy with different thresholds. We find that the overall accuracy is relatively robust to the threshold selection, in the wide range from 0.47 to 0.70.

\begin{table}[tp]
  \footnotesize
\centering
\vspace{-10pt}
\caption{Effect of threshold of confidence score on zero-shot accuracy. }
\scalebox{0.9}{%
\begin{tabular}{c c |c c}
    \toprule
     Threshold & Low conf. set size & Acc on low conf. set & Acc on full set \\
    \midrule
     0.47 & 10000  & 19.40\% & 68.72\%  \\ 
     0.52 & 11000 & 20.82\% & 68.78\%   \\
     0.57 & 12000  & 22.06\% & 68.82\%  \\ 
     0.62 & 13000 &  23.58\% &    68.85\%  \\ 
     0.66 & 14000 &  25.01\% &    \bf{68.88}\%  \\ 
     0.70 & 15000 &  26.51\% &    68.86\%  \\ 
    
     \bottomrule
\end{tabular}
}
\label{table:threshold}
\vspace{-8pt}
\end{table}


\section{Conclusion}
Multi-modal models' generalization and robustness is critical for deployment. 
Motivated by the big gap between top-1 and top-5 accuracy in ImageNet zero-shot classification, we investigated the failure modes and found that the model's prediction is very sensitive to text prompts. 
We describe a simple but efficient zero-shot post-hoc method to identify a subset of samples that are most likely to be predicted wrongly by a measure of self-consistency.
For those in the low confidence subset, we use the WordNet hierarchy to augment class labels to enhance the robustness, resulting in up to 17.13\% accuracy improvement on ImageNet.  
We show our method provides consistent improvement over other distribution shifted datasets (ImageNet variants), four other datasets, and is generalizable to other image-text models and different backbones.

\noindent{\bf Acknowledgments} This work was supported in part by C-BRIC (one of six centers in JUMP, a
Semiconductor Research Corporation (SRC) program sponsored by DARPA),
DARPA (HR00112190134) and the Army Research Office (W911NF2020053). The
authors affirm that the views expressed herein are solely their own, and
do not represent the views of the United States government or any agency
thereof.

{\small
\bibliographystyle{ieee_fullname}
\bibliography{main_arxiv}
}

\clearpage

\appendix


\begin{figure*}[h]
\centering
\includegraphics[width=0.9\textwidth]{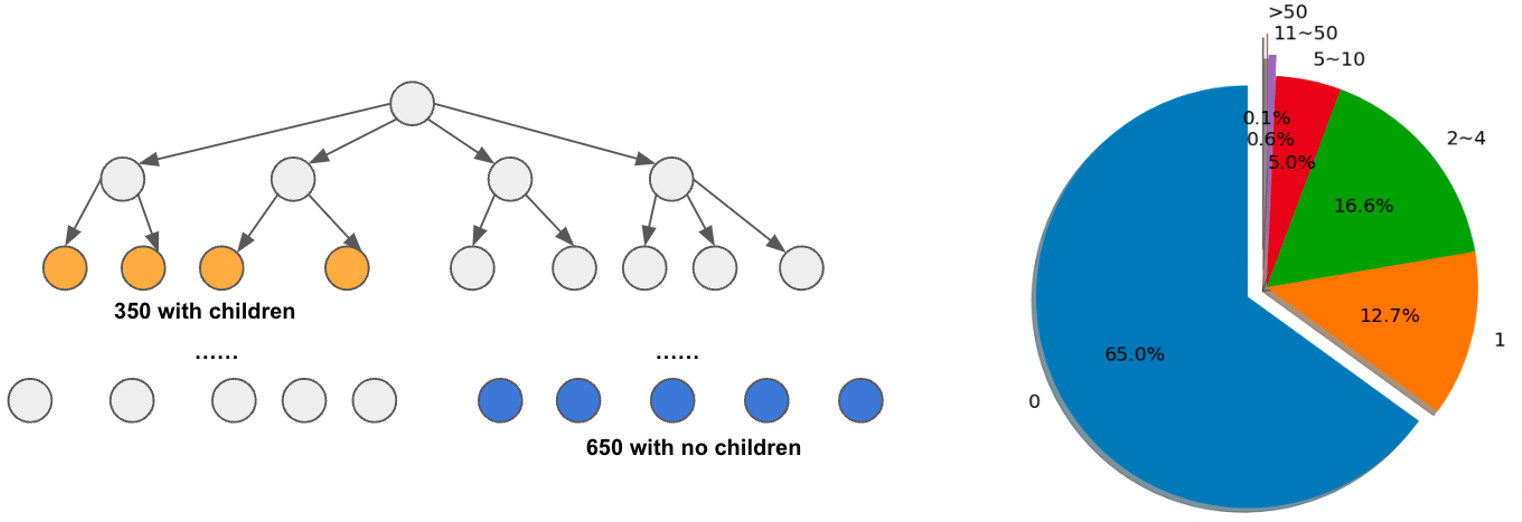}
\caption{ (a) The 1000 ImageNet class names are at different levels of WordNet hierarchy with different degree of abstraction. 350 of them are super-class with sub-classes as the children, while the rest 650 of them have no children. (b) The distribution of the number of children: 12.7\% of the classes have one child node, 16.6\% of the classes have 2-4 child nodes. 
}
\label{fig:hierarchy}
\end{figure*}

\section*{Appendix}

\section{Analyzing the classes for which the top-5 prediction is correct but the top-1 prediction is mostly incorrect}

\begin{table}[htp]
\center
\small
\begin{tabular}{cc}
Ground Truth Class Name                 & Error rate \\ \toprule
tusker                    &      94\%       \\
missile                   &      94\%       \\
terrapin                  &      92\%       \\
collie                    &      90\%         \\
screw                     &     90\%       \\
mushroom                  &  88\%        \\
Appenzeller Sennenhund    &  84\%               \\
snoek fish                &  84\%           \\
husky                     &  82\%               \\
parallel bars             &     82\%           \\
gazelle                   &     82\%        \\
sailboat                  &     82\%        \\
corn cob                  & 80\%        \\
analog clock              & 78\%            \\
cornet                    & 78\%                \\
gossamer-winged butterfly & 76\%            \\
green mamba               & 76\%            \\
tiger cat                 & 74\%            \\
hare                      & 74\%            \\
canoe                     & 72\%          \\ \bottomrule 
\end{tabular}
\caption{List of top 20 classes where the top-5 prediction is correct but the top-1 prediction is mostly incorrect: sorted descendingly by error rate. The \% indicates the proportion of images within the class whose top-5 prediction is correct but whose top-1 prediction is incorrect. }
\label{tab:freq_class_top1_wrong}
\end{table}

\section{Locating the 1000 ImageNet classes at WordNet hierarchy}
Fig.~\ref{fig:hierarchy} shows the location of the 1000 ImageNet classes within the WordNet hierarchy.  The 1000 ImageNet class names are at different levels of WordNet hierarchy with different degrees of abstraction. 350 are super-classes with sub-classes as the children, while the remaining 650 are leaf nodes with no children. (b) The distribution of the number of children: 12.7\% of the classes have one child node and 16.6\% have 2-4 child nodes.

\section{Additional results}\label{app:additional:results}

\begin{figure*}[tp]
\centering
 \includegraphics[width=\textwidth]{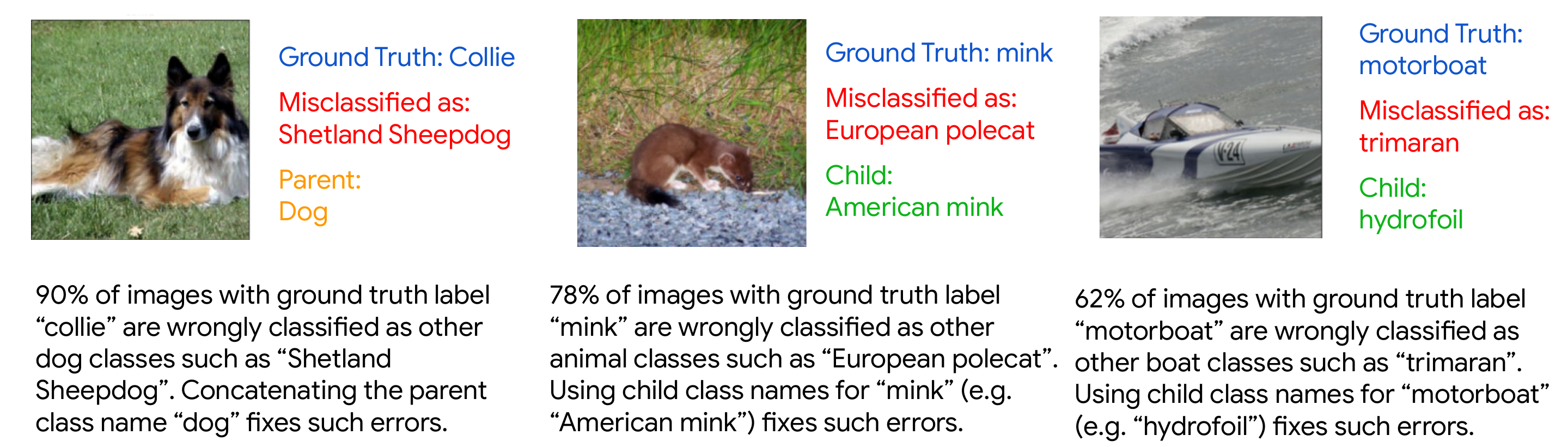}
\caption{ Qualitative visualization on typical failure modes for cases where our top-down and bottom-up prompt augmentation using the WordNet hierarchy method fixes the errors. In each case, the image is originally mis-classified but is correctly classified with our proposed method. 
}
\label{fig:qualitative-result}
\end{figure*}

\paragraph{Qualitative visualization.}
Figure~\ref{fig:qualitative-result} shows a qualitative visualization on more typical failure modes in the cases where our top-down and bottom-up prompt augmentation using the WordNet hierarchy method fixes the error.

\paragraph{Effect of the model architecture and size on selected low-confidence sets on ImageNet.} We found a more powerful backbone leads to a smaller low-confidence sets (e.g., the low-confidence sets of ViT-l14 and ResNet-50 contain 8,557 and 13,106 images, respectively).

\paragraph{Benefits to Top-5 accuracy.} If we apply our top-down and bottom-up label augmentation method to re-rank top-10 classes, we see it can improve the top-5 on the low confidence set from 77.4\% to 80.2\%. We also find reranking top-10 further improves top-1 performance vs. re-ranking top-5 only.

\paragraph{Sparsifying WordNet using the norm of text embedding.} 
WordNet contains many academic words that are rarely used in common usage of English, and hence unlikely to occur frequently in the captions used for CLIP training. For example, ``anthozoan, actinozoan'', ``coelenterate'', ``gastropod'', etc..
Directly using the raw WordNet with academic words as parents is not helpful for improving zero-shot accuracy, and can even hurt the performance. 
Though we do not have access to the CLIP private data, we studied the norm of the word embedding vector and found it is correlated with word frequency. 
We compute the $L_2$ norm of the prompt embedding when plugging in the word into promt templates, i.e., $\rVert f_{text}(t(c)) \rVert$, $t \in \mathcal{T}$. 
We found that the variance of the norm, $\Var_{t\in\mathcal{T}}(\rVert f_{text}(t(c)) \rVert)$, is correlated with word frequency. Rare words tend to have small variances, while common words tend to have large variances. 
For example, the variance of the rare word ``anthozoan'' is 0.118, while the variance of a more common word 
``workplace'' is  0.724. 
We use this statistic to filter out rare words in WordNet. We removed 60\% of the nodes in WordNet and only kept the top 40\% nodes with the highest variance and found this may work slightly better than using the whole WordNet in some cases.
Our intuition behind the correlation between the norm variance and the word frequency is that, for a frequent word that has many examples in the CLIP training data, the CLIP model learns a very precise text embedding such that it has the capability to tell the semantic difference under different contexts, e.g., ``{\small\texttt{a photo of a nice \{label\}}}'' and ``{\small\texttt{a photo of a weird \{label\}}}''.

\begin{table}[tp]
  \footnotesize
\centering
\caption{Effect of WordNet sparsity on zero-shot top-1 accuracy on ImageNet with CLIP.}
\scalebox{1.0}{%
\begin{tabular}{c c c}
    \toprule
     \% of remaining words & acc overall\\
    \midrule
    100\%  &  68.52\%  \\ 
     40\%   &  68.52\%  \\ 
     30\%  &  \bf{68.72}\%   \\
     20\%   &  \bf{68.72}\%  \\ 
     10\%   &  \bf{68.72}\%  \\
     \bottomrule
\end{tabular}
}
\label{table:threshold-wordnet}
\vspace{-12pt}
\end{table}

\paragraph{Effect of WordNet sparsity on zero-shot accuracy}
We evaluate the effect of the degree of sparsity of WordNet on the downstream zero-shot accuracy. 
We sparsify the WordNet based on word frequency, which is measured by embedding variance as described in the previous section. 
Here we study the effect of sparsity on the downstream zero-shot accuracy. 
Table~\ref{table:threshold-wordnet} shows the overall accuracy on ImageNet using CLIP model with different levels of WordNet sparsities  .

\end{document}